\def\year{2022}\relax
\documentclass[letterpaper]{article} % DO NOT CHANGE THIS
\usepackage{aaai22}  % DO NOT CHANGE THIS
\usepackage{times}  % DO NOT CHANGE THIS
\usepackage{helvet}  % DO NOT CHANGE THIS
\usepackage{courier}  % DO NOT CHANGE THIS
\usepackage[hyphens]{url}  % DO NOT CHANGE THIS
\usepackage{graphicx} % DO NOT CHANGE THIS
\usepackage{natbib}  % DO NOT CHANGE THIS AND DO NOT ADD ANY OPTIONS TO IT
\usepackage{caption} % DO NOT CHANGE THIS AND DO NOT ADD ANY OPTIONS TO IT
\usepackage{multirow}
\usepackage{subcaption}
\DeclareCaptionStyle{ruled}{labelfont=normalfont,labelsep=colon,strut=off} % DO NOT CHANGE THIS
\usepackage{algorithm}
\usepackage{algorithmic}

\usepackage{newfloat}
\usepackage{listings}
\floatstyle{ruled}
\newfloat{listing}{tb}{lst}{}
\floatname{listing}{Listing}
\title{Fusing Task-oriented and Open-domain Dialogues in Conversational Agents}
\author{
    Tom Young, \textsuperscript{\rm 1}
    Frank Xing, \textsuperscript{\rm 2}
    Vlad Pandelea, \textsuperscript{\rm 1}
    Jinjie Ni, \textsuperscript{\rm 1}
    Erik Cambria\textsuperscript{\rm 1}
}
\affiliations{
    \textsuperscript{\rm 1} School of Computer Science and Engineering, Nanyang Technological University, Singapore\\
    \textsuperscript{\rm 2} School of Computing, National University of Singapore, Singapore\\
    yang0552@e.ntu.edu.sg, xing@nus.edu.sg, \{vlad.pandelea, jinjie001\}@e.ntu.edu.sg, cambria@ntu.edu.sg
}

\begin{document}

\maketitle

\begin{abstract}
The goal of building intelligent dialogue systems has largely been \textit{separately} pursued under two paradigms: task-oriented dialogue (TOD) systems, which perform task-specific functions, and open-domain dialogue (ODD) systems, which focus on non-goal-oriented chitchat. The two dialogue modes can potentially be intertwined together seamlessly in the same conversation, as easily done by a friendly human assistant. Such ability is desirable in conversational agents, as the integration makes them more accessible and useful. Our paper addresses this problem of fusing TODs and ODDs in multi-turn dialogues. Based on the popular TOD dataset MultiWOZ, we build a new dataset FusedChat, by rewriting the existing TOD turns and adding new ODD turns. This procedure constructs conversation sessions containing exchanges from both dialogue modes. It features inter-mode contextual dependency, i.e., the dialogue turns from the two modes depend on each other. Rich dependency patterns such as co-reference and ellipsis are included. The new dataset, with 60k new human-written ODD turns and 5k re-written TOD turns, offers a benchmark to test a dialogue model's ability to perform inter-mode conversations. This is a more challenging task since the model has to determine the appropriate dialogue mode and generate the response based on the inter-mode context. However, such models would better mimic human-level conversation capabilities. We evaluate two baseline models on this task, including the \textit{classification-based} two-stage models and the \textit{two-in-one} fused models. We publicly release FusedChat and the baselines to propel future work on inter-mode dialogue systems.

\end{abstract}

\section{Introduction}

Recent years have seen a popularity of models on building intelligent systems that converse with humans naturally~\cite{ni2021recent}. Two mainstream models can be categorized as the open-domain dialogue (ODD) models~\cite{adiwardana2020towards,roller2020recipes,zhang2019dialogpt} and the task-oriented dialogue (TOD) models~\cite{ham2020end,budzianowski2018multiwoz}.
ODD models, when first adapted with the Seq2Seq modeling paradigm~\cite{sutskever2014sequence}, focused on learning open-domain human conversation based on massive [context, response] pairs~\cite{Vinyals2015A,li2015diversityMMI}. 

\begin{figure}[h]
\includegraphics[width=\linewidth]{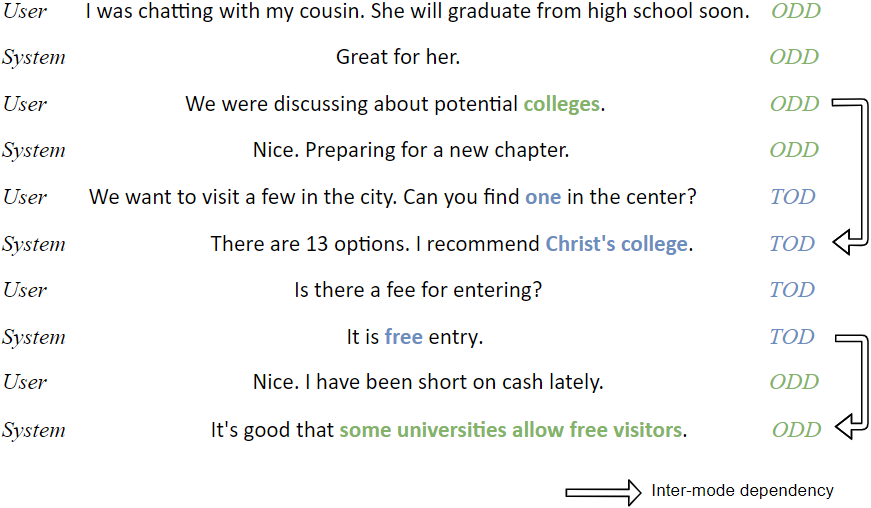}
\caption{Example of interaction with our dialogue system. The conversation between a user and a digital assistant seamlessly interchanges between TOD and ODD modes with strong inter-mode dependency. The conversation involves querying about a college entrance fee (TOD) and chitchat about personal development and finance (ODD).}
\label{fig:toy_example}
\end{figure}

Such models generate the response based on the context and exhibit general chitchat ability. Their primary goal in a conversation is to keep the user engaged and chat over random open-domain topics that he is interested in. The dialogues can be sustained by commonsense without the need for any special databases. 
TOD models are vastly different. The dialogues exist for the purpose of serving specific functions, such as finding restaurants and booking airlines. They operate on closed domains that are often supported by structured databases and APIs~\cite{budzianowski2018multiwoz,rastogi2020towards}. Commonly three characteristics distinguish them from ODD models: (1) an entity-centered database, (2) explicit dialogue state modeling, and (3) a pre-defined set of dialogue domains and functions (dialogue acts).
Humans are able to effortlessly conduct both types of conversations seamlessly together. It is ideal for a dialogue system to be able to do so, because such integration offers a fused system with increased usability. Furthermore, it allows rich interactions between the two dialogue modes, which can not be modeled in either mode independently. Such a dialogue model would better mimic human-level conversation capabilities, e.g., chatting with a friendly assistant (Fig.~\ref{fig:toy_example}).

Despite numerous datasets have been created in recent years for both ODDs and TODs, there is no high-quality human-written dataset on their fusion, especially with inter-mode contextual dependency. Our work aims to fill this void. We use the popular TOD dataset MultiWOZ~\cite{budzianowski2018multiwoz} as the backbone and let human creators add ODD turns before or after the existing TOD turns. For roughly half the MultiWOZ dialogues, we prepend ODD turns, creating ODD + TOD sessions. For the other half, we append ODD turns, creating TOD + ODD sessions. In both cases, the creator writes an ODD that is contextually related to the existing TOD. We enforce inter-mode dependency in FusedChat. In the prepending case, we make sure the TOD is dependent on the ODD by rewriting the first turn of the TOD, typically with co-reference or ellipsis. In the appending cases, we make sure at least one exchange in the ODD is dependent on concepts or knowledge found in the TOD. In a nutshell, these dependency patterns in our dataset mean that when a dialogue model handles a turn of one dialogue mode, it sometimes has to refer to the contextual information given in the history turns of the other dialogue mode.\\

This new dataset offers a unique test-bed for training and evaluating inter-mode dialogue systems that possess both TOD and ODD capabilities. Traditional dialogue evaluation metrics for both dialogue modes can be used together for inter-mode evaluation. 

We develop and evaluate two baseline models for this new setting: (1) The \textit{classification-based} model. Two response generation models $\mathcal{M}_{tod}$ and $\mathcal{M}_{odd}$ are independently trained on the turns of the respective modes. They generate the response of their respective mode given a conversational context. A separate mode classification model $\mathcal{C}$ is trained and used to determine which mode to invoke given the context. (2) The \textit{two-in-one} fused dialogue model that is trained on dialogue turns of both modes together. It generates a response given any conversational context, by implicitly predicting the dialogue mode as part of sequence generation.

In summary, our main contributions are: (1) A new dialogue dataset named FusedChat\footnote{\url{https://github.com/tomyoung903/FusedChat}} that fuses TODs and ODDs in multi-turn dialogues. The dialogues feature inter-mode contextual dependency for seamless mode fusion, allowing the dialogue model to better mimic human-level conversation capabilities. FusedChat, with 60k new human-written ODD turns and 5k re-written TOD turns, serves as a new benchmark for inter-mode dialogue systems. Traditional metrics used to gauge TOD and ODD systems separately can be combined to evaluate inter-mode dialogue systems. (2)  \textit{two-in-one} models and \textit{classification-based} models are developed and evaluated as inter-mode dialogue models. Our preliminary experiments suggest that the models perform worse than their single-mode counterparts evaluated on single-mode datasets. And the more computationally expensive \textit{classification-based} model outperforms the cheaper \textit{two-in-one} fused model. This suggests that effectively fusing different dialogue modes is a challenging task and there is a huge room for improvement upon our baseline fusion models. 

\section{FusedChat Construction}

To create inter-mode dialogue sessions, our dataset construction process mainly involves having dialogue creators prepend or append self-written ODDs to existing TODs. A dialogue creator plays the part of both the user and the dialogue system by himself. This self-dialogue setting~\cite{byrne2019taskmaster} avoids misunderstandings between two human creators and improve the consistency of the created dialogues.

For the existing TODs, the MultiWOZ 2.4 dataset~\cite{ye2021multiwoz} is selected because of its popularity in the literature. MultiWOZ contains TODs in 7 domains, including restaurant, attraction, train, police, hospital, taxi and hotel. The user converses with the dialogue agent for a pre-defined set of functions, such as booking restaurants and locating hospitals. Despite that MultiWOZ was created assuming the user is a tourist~\cite{budzianowski2018multiwoz}, we find that most dialogues themselves do not necessarily reflect a tourist persona and allow flexibly adding open-domain dialogues. In our FusedChat setting, the dialogue creators are free to add any ODD that is contextually consistent with the existing TOD.

In the following sections, we first discuss the general requirement we set for the added ODDs. We then explain how prepending and appending ODDs are executed and how inter-mode dependency is enforced, respectively.

\subsection{General requirements for the added ODDs} \label{sec:appending_inter_mode}

In this section, we describe the general requirements for the added ODDs for both the prepending and appending cases, as rules for the dialogue creators to follow.

(1) Every creator writes fictitious ODDs for \textit{both} the roles of ``system'' and ``user'', where the ``system'' represents an AI conversational agent that is capable of both friendly open-domain conversation (in the added ODDs) and task-oriented dialogues (in the existing MultiWOZ TODs). And ``user'' represents a human speaker that converses with the AI agent for friendly chitchat and to achieve certain task objectives. 

(2) To ensure the relevance between the existing TOD and the added ODD, we encourage the creators to make the ODD revolve around similar or related topics as in the existing TOD segment, e.g., by talking about the same or related concepts in the TOD. The added ODD turns and the existing TOD turns should connect with each other naturally. There should be strong contextual dependency between the two modes (explained in the next 2 sections).

(3) The created dialogues should adhere to the general characteristics of ODDs as opposed to TODs. They should be casual chitchat exchanges that do not require the ``system'' to perform any specific task-oriented functionalities or provide any task-specific information.

\begin{itemize}
  \item Based on the pilot experiment with a sample of creators, we found that the creators had a tendency to write dialogues that are focused on task-specific functionalities, which are technically TODs instead of ODDs as instructed. This is presumably because of a lack of nuanced understanding of their difference, and the ease of fitting those TODs into the context of existing TODs.\\As an aggressive measure to combat this issue, we deployed a real-time turn-level ODD vs TOD classifier, trained on a combination of three traditional ODD datasets~\cite{zhang2018personalizing, smith2020can, dinan2018wizard} and MultiWOZ. In addition, we outline several pitfalls found in the pilot experiment for the creators to avoid, such as letting the system fabricate information that is beyond commonsense.
\end{itemize}

Next, we describe the details on how appending ODDs (TOD + ODD) and prepending ODDs (ODD + TOD) are executed, and how inter-mode dependency is enforced, respectively.

$ $\\

\subsection{Appending ODDs}
In the appending scenario, the dialogue creators append an ODD to a provided TOD sampled from the MultiWOZ dataset. The ODD should naturally \textit{follow} the TOD.

\begin{itemize}
\item
We notice that the dialogues from the original MultiWOZ dataset often end with a ``User: Thank you. System: Goodbye.'' exchange. This exchange effectively \textit{ends} the conversation. For appending ODDs, we heuristically remove such exchanges from the end of the TOD based on dialogue act annotations (dialogue-act:thank-you and dialogue-act:goodbye).
\end{itemize}

\subsubsection{Inter-mode Dependency} \label{sec:appending_inter_mode}

\begin{figure}[t]
\includegraphics[width=\linewidth]{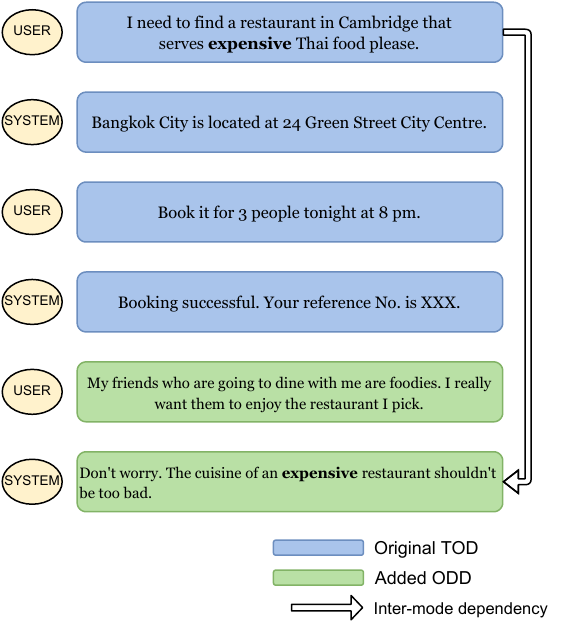}
\caption{An excerpt from a TOD + ODD instance from FusedChat. Note how inter-mode dependency is featured in the last system ODD turn by referring to the concept ``expensive restaurant'' previously mentioned in the TOD.}
\label{fig:dependency_appending}
\end{figure}

In appending cases, the content of the ODD should be dependent on the preceding TOD. We enforce this by letting the creators write at least one round of exchange whose content reflects concepts or knowledge found the existing TOD segment. %, without any clues from the previously added chitchat user turns (i.e., if the concepts reflected happen to exist in a previously added chitchat user turn, then it doesn't count). 
Fig.~\ref{fig:dependency_appending} shows a TOD + ODD example. The first two rounds of exchange between the user and the system is under the TOD mode. They are about querying and booking an expensive Thai restaurant. The system's replies are supported by dialogue state tracking~\cite{budzianowski2018multiwoz} and an underlying database on available restaurants. In the third round of exchange, the user expresses concern over whether his friends would enjoy the restaurant. Note that this is considered an ODD utterance since it does not invoke any task-oriented function. The system's ODD response is supported by commonsense and empathy. Note how it reflects content from a history TOD turn.

\subsection{Prepending ODDs}

In prepending cases, the creator is given a TOD segment from MultiWOZ and asked to prepend an ODD to it. The ODD should naturally \textit{lead to} the provided TOD.

Note that the original TODs in MultiWOZ are self-contained. For our purpose of modeling inter-mode dependency, we conduct utterance rewriting based on co-reference and ellipsis. In FusedChat, they are the key why the TOD is dependent on the prepended ODD.

\subsubsection{Inter-mode Dependency}

We want to create ODD + TOD sessions where the TOD is conditioned on the ODD. The key to a successful TOD is dialogue state tracking, where the dialogue system processes the user utterance for [slot type, slot value] pairs (e.g., [Destination: Cambridge]) in order to understand the user's need and respond properly. Our designed method to model inter-mode dependency in our dataset essentially imposes ODD-dependent dialogue state tracking.

\begin{figure}
  \centering
  \begin{subfigure}[b]{\linewidth}
    \includegraphics[width=0.9\textwidth]{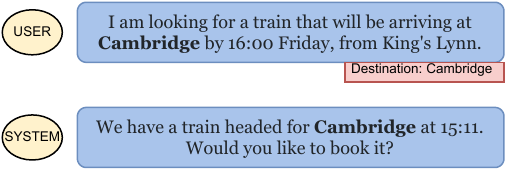}
    \caption{An original TOD exchange with the dialogue state [Destination: Cambridge].}
    \label{fig:original_tod}
  \end{subfigure}
  
  \vspace{0.05\textwidth}%

  \begin{subfigure}[b]{\linewidth}
    \includegraphics[width=1\textwidth]{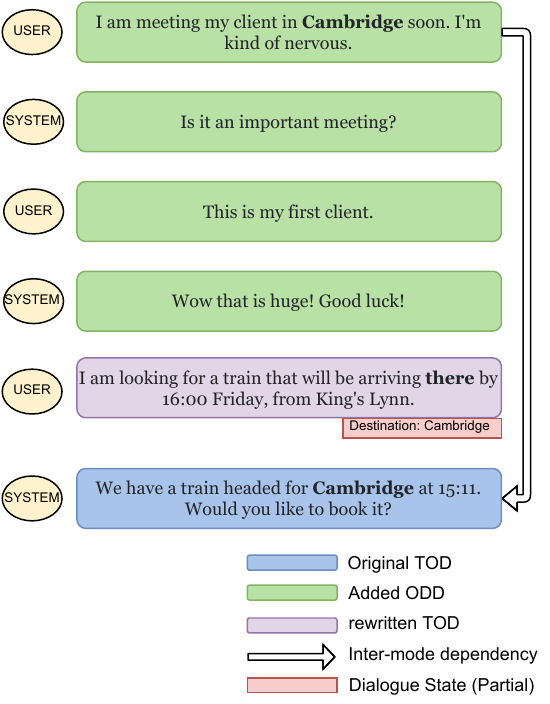}
    \caption{New ODD turns are prepended to the original TOD in (a). Note that the TOD user turn is rewritten. The slot value ``Cambridge'' is mentioned in a prepended ODD turn while co-reference is used in the rewritten user turn. This imposes ODD-dependent dialogue state tracking, forcing the the dialogue system to look for clues in the ODD when it tries to interpret the user's need. }
    % A seamless inter-mode dialogue session is created in this way, where the following TOD turn is dependent on added TOD.
    \label{fig:}
  \end{subfigure}

\caption{An ODD + TOD instance from FusedChat.}
\label{fig:dependency_prepending}
\end{figure}

We randomly select a slot value mentioned in the first user turn in the TOD, e.g., ``Cambridge'' in Fig.~\ref{fig:dependency_prepending}. We ask the dialogue creators to use the slot value in the prepended ODD, and rewrite the first dialogue user turn accordingly to refer to it implicitly. Rewriting mainly involves co-reference (e.g., ``there'' in Fig.~\ref{fig:dependency_prepending}), and sometimes ellipsis. Co-reference and ellipsis are important features in multi-turn TODs, attracting researchers to sometimes perform special annotations for them in certain TOD datasets~\cite{quan2020risawoz}. See Fig.~\ref{fig:dependency_prepending} for a detailed example on how inter-mode dependency is featured for ODD + TOD sessions.

$ $

\section{FusedChat statistics}

A total of 113 undergraduate students from the authors' university were recruited as dialogue creators for FusedChat. The difference between FusedChat and MultiWOZ mainly lies in the additional ODD turns, either grounding or grounded by the original TODs.
The added ODD turns in FusedChat are a significant extension to the original MultiWOZ dataset. As shown in Table~\ref{tab:statistics}, 60k+ new ODD turns are added, including 8k+ new tokens not present in the original MultiWOZ dataset, significantly expanding the vocabulary.

FusedChat also rewrote the first TOD turns (4670 in total) for the scenario of prepending ODDs. For the scenario appending ODDs, FusedChat discarded 11320 TOD turns containing only ``thank-you'' and ``goodbye'' dialogue acts. Table~\ref{tab:partition} shows the training/validation/testing partition for FusedChat.

\begin{table*}[]
\centering
\begin{tabular}{|c|c|}
\hline
Total No. turns                                                         & 60579  \\ \hline
Total No. tokens                                               & 680448 \\ \hline
Avg. No. turns per dialogue                                       & 5.81   \\ \hline
Avg. No. tokens per turn                                 & 11.23  \\ \hline
No. unique tokens                                   & 11822  \\ \hline
No. unique tokens not present in MultiWOZ & 8075   \\ \hline
\end{tabular}
\caption{Statistics on the added ODD turns in FusedChat}
\label{tab:statistics}
\end{table*}

\begin{table*}[]
\centering
\begin{tabular}{|c|c|c|c|}
\hline
Partition   & ODD + TOD &  TOD + ODD &   Total \\ \hline
Training    & 3670      &      4768 &      8438         \\ \hline
Validation  & 500       &      500  &      1000           \\ \hline
Testing     & 500       &      500  &      1000          \\ \hline
Total       & 4670      &     5768  &      10438         \\ \hline
% Domains/slot names     & 519     &  481 &      523       \\ \hline
\end{tabular}
\caption{FusedChat is composed of ODD + TOD (prepending ODDs) instances and TOD + ODD (appending ODDs) instances.}
\label{tab:partition}
\end{table*}

$ $

\section{Approaches for inter-mode dialogues}\label{sec:approaches}

In this section, we discuss baseline models we developed for inter-mode dialogues.

\subsection{Task Definition}
A multi-turn dialogue system generates a response $R$ based on a multi-turn context $C$. In inter-mode dialogues, $C$ is composed of both TOD and ODD turns. In the FusedChat setting, $R$ can be in either TOD mode or ODD mode, but has to be in only one of the two.\\

\subsection{Models} \label{sec:models}
We experiment with two types of models for inter-mode dialogues. (1) The \textit{classification-based} model that is composed of a mode classification model and two response generation models for TOD and ODD separately and (2) the \textit{two-in-one} fused model where a single response generation model can perform both TOD and ODD generation, implicitly determining the response mode.

(1) The \textit{classification-based} model. Two response generation models $\mathcal{M}_{odd}$ and $\mathcal{M}_{tod}$ are independently trained to handle each conversation mode. A separate classification model $\mathcal{C}$ is trained and used to determine which mode of model to invoke given an inter-mode context. Note that all 3 models above take inter-mode context as input.

\begin{itemize}

\item For $\mathcal{M}_{odd}$, we follow~\cite{shuster2019dialogue} and experiment with DialoGPT~\cite{zhang2019dialogpt} as the pretrained model, fine-tuned on all ODD turns in FusedChat. 
% BlenderBot, Plato-2, and Meena, gpt-3

\item For $\mathcal{M}_{tod}$, we follow the recent progress on end-to-end modeling for TODs. Dialogue state tracking, dialogue act prediction and response generation have been together cast under a Seq2Seq framework~\cite{hosseini2020simple, ham2020end}. For traditional Seq2Seq-based ODD modeling, the problem is cast as [Context $\rightarrow$ Response]. For Seq2Seq-based TOD modeling, the problem is cast as [Context $\rightarrow$ (Dialogue State, Dialogue Act, Response)], where the three latter parts are concatenated together as one sequence as the generation target. This simplistic form allows TOD models to exploit the benefits of large-scale pretrained models, same as ODD models did. We follow \textit{Neural Pipeline}~\cite{ham2020end} for such a model for $\mathcal{M}_{tod}$, initialized with GPT2 and fine-tuned on all TOD turns in FusedChat.

\item For $\mathcal{C}$, we follow~\cite{madotto2020adapter} and experiment with BERT~\cite{devlin2018bert} as the pretrained model. The model is fine-tuned on all turns in FusedChat to predict the dialogue mode (TOD vs ODD).

\end{itemize}

(2) The \textit{two-in-one} model. Trained on dialogue turns of both modes, it uses a single model that generates a response given any conversational context by implicitly determining the conversational mode. Similar to~\cite{sun2020adding}, we use an additional $<$ODD$>$ token during sequence construction to indicate when the response is in the ODD mode. The training sequences are composed of [Context $\rightarrow$ ($<$ODD$>$, Response)] and [Context $\rightarrow$ (Dialogue State, Dialogue Act, Response)]. The model is initialized with GPT2 and fine-tuned on all dialogue turns in FusedChat.

For all the models above, best checkpoints for testing are selected based on the full validation sets of 1000 instances.

\section{FusedChat as a new benchmark}\label{sec:fusedchat_benchmark}

\begin{figure}[h]
\includegraphics[width=\linewidth]{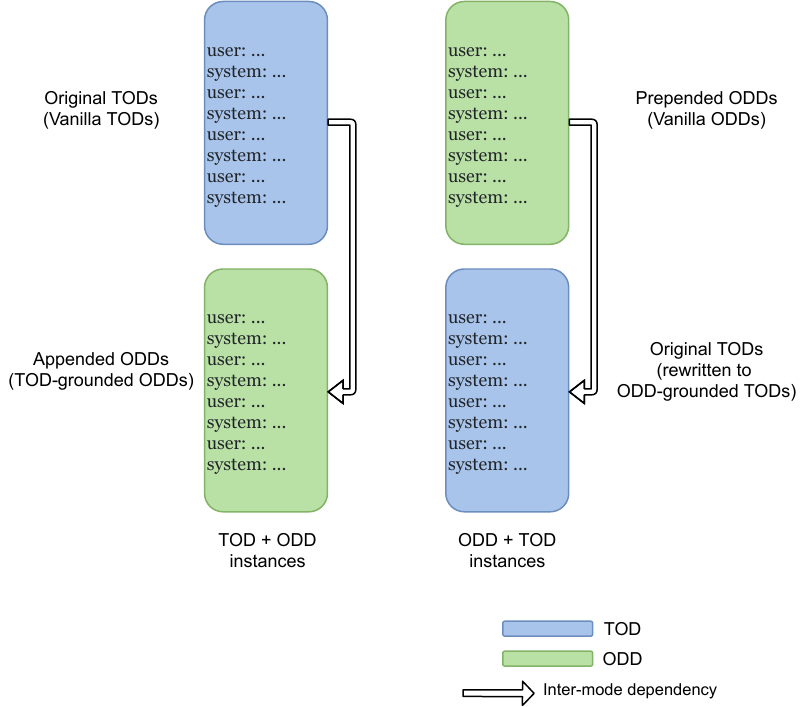}
\caption{4 types of dialogue turns are present in FusedChat, classified by the dialogue mode and the grounding context.}
\label{fig:four_types}
\end{figure}

Depending on the context and the dialogue mode, the dialogue turns in our dataset are naturally separated into 4 types in Fig.~\ref{fig:four_types}: vanilla TODs, vanilla ODDs, ODD-grounded TODs and TOD-grounded ODDs. Vanilla refers to the dialogue turns being grounded on context of its own mode only, resembling traditional datasets. The ODD turns in the ``prepending ODDs'' scenario and TOD turns in the ``appending ODDs'' scenario are vanilla.

In the following sections, we illustrate 4 unique evaluation scenarios on which FusedChat can benchmark the performance of inter-mode dialogue systems, including mode classification, TOD-grounded ODDs, ODD-grounded TODs and full inter-mode dialogues.

\subsection{Mode classification}
The straightforward problem in inter-mode dialogues is to decide which mode the generated response should be. Should the system respond with friendly chitchat (ODD), or should it try to interpret the user's task-oriented goal and respond with certain dialogue acts (TOD)? The accuracy for the mode classification model is shown in Table~\ref{tab:mode_class_acc}. We consider two context options: using only the latest user turn as the context (single-turn) or using the whole history containing multiple turns as the context (multi-turn). Results show that the accuracy is quite high in both cases, with ``multi-turn'' marginally outperforming ``single-turn''.

$ $

\begin{table}[h]
\centering
\begin{tabular}{|c|c|}
\hline
Context option                    & Accuracy  \\ \hline
Single-turn                     & 0.993 \\ \hline
Multi-turn                      & 0.995   \\ \hline
\end{tabular}
\caption{Mode classification accuracy for model $\mathcal{C}$.}
\label{tab:mode_class_acc}
\end{table}

\begin{table*}[h]
\centering
\begin{tabular}{|c|c|c|c|c|c|}
\hline
Models               & Slot Accuracy (SA) & Joint SA & Inform & Success & BLEU  \\ \hline
\multicolumn{6}{|c|}{\textbf{ODD-grounded TODs in FusedChat}}                                            \\ \hline
\textit{Two-in-one} model          & 0.971         & 0.574               & 71.1        & 56.9         & 12.16 \\ \hline
\textit{Classification-based} model & 0.972         & 0.584               & 72.8        & 60.0         & 12.58 \\ \hline
\multicolumn{6}{|c|}{\textbf{Original MultiWOZ dataset}}                                                 \\ \hline
\textit{Neural Pipeline}~\cite{ham2020end}      & 0.976         & 0.631               & 79.2        & 64.3         & 12.72 \\ \hline
\end{tabular}
\caption{Evaluation results on ODD-grounded TODs in FusedChat and comparison with MultiWOZ results.}
\label{tab:odd-g tod}
\end{table*}

\begin{table*}[h]
\centering
\begin{tabular}{|c|c|c|c|c|}
\hline
Models & PPL & Sensibleness & Specificity  & SSA \\ \hline
% $\mathcal{M}_{odd}$  &              &       &       &  \\ \hline
\textit{Two-in-one} model  & 9.15 &     0.44         &    0.39  & 0.42 \\ \hline
\textit{Classification-based} model & 8.79 &    0.51       & 0.45  &  0.48 \\ \hline
%  adapter-bot &               &       &         &  \\ \hline
 Ground-truth & \textit{N/A} &    0.97          &  0.91            &  0.94 \\ \hline
% 9.58
\end{tabular}
\caption{Evaluation results on TOD-grounded ODDs in FusedChat.}
\label{tab:tod-g odd}
\end{table*}

\begin{table*}[!htbp]
\centering
\begin{tabular}{|c|c|c|c|c|c|c|c|c|c|}
\hline
\multirow{2}{*}{Models}                                                        & \multicolumn{5}{c|}{TOD metrics}                    & \multicolumn{4}{c|}{ODD metrics}          \\ \cline{2-10} 
                                                                               & Slot Accuracy & Joint SA & Inform & Success & BLEU  & PPL   & Sensibleness & Specificity & SSA  \\ \hline
\begin{tabular}[c]{@{}c@{}}\textit{Two-in-one} \\ model\end{tabular}           & 0.972         & 0.592    & 70.4   & 57.0    & 12.05 & 10.49 & 0.52         & 0.47        & 0.50 \\ \hline
\begin{tabular}[c]{@{}c@{}}\textit{Classification-based} \\ model\end{tabular} & 0.973         & 0.600    & 75.1   & 60.9    & 12.17 & 10.50  & 0.58         & 0.51        & 0.55 \\ \hline
\end{tabular}
\caption{Evaluation results on the full FusedChat testset}
\label{tab:fusedchat}
\end{table*}

\subsection{ODD-grounded TODs}

Part of inter-mode dialogues are ODD-grounded TODs, which correspond to the ``prepending ODDs'' scenario in FusedChat. Like in regular TODs, the system's response is prompted by a task-oriented user request. However, the preceding context contains ODD exchanges, which create unique challenges. 

On the one hand, the model needs to recognize useful task-related information from the ODD context for correct dialogue state tracking. On the other hand, the system's response should correctly perform the required task-oriented function according to the latest user request, instead of derailing to chitchat by following the ODD context in the history.

Evaluation results for this portion of the dialogue turns in FusedChat are shown in Table~\ref{tab:odd-g tod}. We use the traditional TOD evaluation metrics for MultiWOZ, where slot accuracy measures dialogue state tracking, inform rate and success rate measure user goal success and BLEU measures response quality (see more details in~\cite{budzianowski2018multiwoz}).

In addition, we evaluate the \textit{Neural Pipeline} approach using the original MultiWOZ dataset (trained and tested on MultiWOZ). Remember that the \textit{classification-based} model contains $\mathcal{M}_{tod}$, which exactly follows the \textit{Neural Pipeline} model. This is to evaluate the difficulty of the new ODD-grounded TOD task compared with the vanilla TOD task in MultiWOZ. Table~\ref{tab:odd-g tod} shows that:

(1) the \textit{classification-based} model outperforms the \textit{two-in-one} model marginally.

(2) The \textit{Neural Pipeline} model evaluated on the same vanilla TOD dialogues in MultiWOZ significantly outperforms the \textit{classification-based} model evaluated on ODD-grounded TODs in FusedChat. Such significant difference suggests that ODD-grounded TODs are a more challenging task than vanilla TODs. Presumably, this is because (a) the extra requirement to correctly determine the response mode and (b) the extra need for ODD-dependent dialogue state tracking.

\subsection{TOD-grounded ODDs}

Another part of inter-mode dialogues are TOD-grounded OODs, which correspond to the ``appending ODDs'' scenario in FusedChat. The system's ODD response should be conditioned on both the TOD and ODD turns in the context.

The evaluation on open-domain dialogue generation is notoriously difficult and numerous evaluation methods have been proposed~\cite{ni2021recent}. In our experiment, we follow~\cite{adiwardana2020towards} and use perplexity plus sensibleness and specificity average (SSA) as metrics. SSA represents the average between sensibleness (\textit{Does the response make sense given the context?}) and specificity (\textit{Is the response specific to the context?}). Both of them are binary for each response. A response can only be deemed specific if it is deemed sensible. SSA results are computed by averaging 5 expert human evaluators' judgement on 100 randomly sampled dialogue turns from the testset. Table~\ref{tab:tod-g odd} shows the performance of the inter-mode dialogue models on this task. 

The \textit{classification-based} model outperforms the \textit{two-in-one} model marginally. Results also show that ground-truth responses receive very high SSA scores, significantly exceeding the better dialogue model of the two we developed. This suggests that there is huge room for improvement on this task.

\subsection{Full inter-mode dialogues}
We show the results on the full FusedChat testset (containing all 4 types of dialogue turns) in Table~\ref{tab:fusedchat}. A combination of TOD and ODD metrics discussed above can be used to holistically gauge a dialogue system's capability to perform inter-mode dialogues. The
\textit{classification-based} model marginally outperforms the \textit{two-in-one} model.

% This seems to suggest that the current baseline \textit{two-in-one} fusion model, although much more computationally efficient, does not capture both dialogue modes simultaneously as well as the \textit{classification-based} pipelined model does. 
Note that for the evaluation of ODD-grounded TODs, TOD-grounded ODDs and full inter-mode dialogues, we evaluate the response in a mode-tolerant manner. This means that even when the model generates a response of the wrong mode, we still evaluate that instance normally, instead of directly punishing the metric value to 0. For example, when evaluating BLEU, we still normally calculates the BLEU score against the ground-truth response even if the response generated by the inter-mode dialogue model is an ODD response. Of course, getting the mode wrong typically means bad scores.

\section{Related Work}

Recently, there have been multiple efforts on developing dialogue systems multi-tasking on various types of dialogues~\cite{ni2021recent,maasur}. Adapter-Bot~\cite{madotto2020adapter} uses a fixed backbone conversational model (DialoGPT) and triggers on-demand dialogue skills (e.g., empathetic responses, weather information, movie recommendation) via different adapters~\cite{houlsby2019parameter}. ~\cite{madotto2020attention} learns a dialogue system that independently parameterizes different dialogue skills, and learns to select and combine each of them through Attention over Parameters. \citeauthor{xuuend} (\citeyear{xuuend}) proposed an end-to-end dialogue model based on a hierarchical encoder-decoder, which employed a discrete latent variable to learn underlying dialogue intentions. They 
argued that the latent discrete variable interprets the intentions that guide machine responses generation~\cite{howint}. \citeauthor{shuster2019dialogue} (\citeyear{shuster2019dialogue}) multi-tasked on 12 separate dialogue datasets that focus on different skills and showed that a single unified model can perform decently well on all tasks. However, these models do not model the dependency between different types of dialogues in the multi-turn setting. Thus, they are not guaranteed to converse seamlessly and naturally in multiple dialogue modes simultaneously in a multi-turn conversation session.

Unlike the models trained on separate dialogue datasets,~\citeauthor{smith2020can} (\citeyear{smith2020can}) tried to fuse multiple skills into one conversation session. They built a new dialogue dataset named Blendedskilltalk containing dialogues where knowledge, emotional and personalizing skills are shown together in the same multi-turn conversation. They show that systems fine-tuned on the new multi-skill dataset have improved ability in handling multiple skills simultaneously in the same multi-turn conversation session. However, they only target open-domain conversations. Our work, on the other hand, targets the fusion of general ODDs and TODs, as we view them as the two most mainstream forms of dialogues for the research community currently. Along the direction of fusing TODs and ODDs,~\citeauthor{zhao2017generative} (\citeyear{zhao2017generative}) proposed to artificially augment task-oriented dialogues with randomly sampled utterances from a chitchat corpus, mainly to improve the out-of-domain recovery performance for the TOD system.

\citeauthor{sun2020adding} (\citeyear{sun2020adding}) proposed to decorate TOD responses with ODD snippets, in order to make the dialogue agent sound more engaging and interactive. Unlike~\cite{sun2020adding}, where ODD snippets act as a supplementary role to TOD responses, our dataset tackles the fusion of TODs and ODDs by treating them as parallel dialogue modes of equal importance, and focuses on modeling inter-mode dependency in the multi-turn setting.

\section{Discussions and Future Work}

Our work serves the goal to develop dialogue systems that are capable of performing both TODs and ODDs with inter-mode dependency. Compared with traditional datasets, the new dataset FusedChat uniquely contains ODD-grounded TODs and TOD-grounded ODDs. It endeavors to fuse the two common forms of human conversations, i.e., casual open-ended conversations supported only by commonsense, and task-oriented conversations supported by specific knowledge bases. We show preliminary experiment results on two baseline models, which suggest huge room for improvement. We release dataset and baselines in order to propel future work on inter-mode dialogue systems.

We note that the framework set by FusedChat is limited. The dataset does not contain dialogue sessions containing more than one mode switch, which represents a gap with real-world scenarios. We suspect more mode switches could make inter-mode dialogues even more challenging. Our choice of TODs and ODDs does not represent the full scope of possible dialogue settings. We chose the most simple form of ODDs where the response is only determined by the context. Yet in the literature, ODDs have been grounded on various forms of information, such as personas~\cite{zhang2018personalizing}. We chose the classical setting of TODs as in MultiWOZ, which is defined by structured entity-centric knowledge bases. However, the concept of TODs has seen expansion, such as with unstructured knowledge access~\cite{kim2020beyond}. We expect the fusion of more complex forms of ODDs and TODs to be more challenging, but they would even better represent human-level conversational abilities.

The construction of FusedChat required a lot of manual creative effort. It is thus very expensive to replicate the same routine for every new inter-mode dialogue scenario. Alternatively, zero-shot or few-shot models that can learn to perform inter-mode dialogues by mostly relying on separate single-mode dialogues are a promising direction. FusedChat can also serve as a test-bed for such paradigms.

$ $

\section*{Acknowledgments}

This research is supported by the Agency for Science, Technology and Research (A*STAR) under its AME Programmatic Funding Scheme (Project \#A18A2b0046). We thank Lu Cheng for creating the data collection interface and Low Shi Min and Arya Shashwat for quality control. We thank Peter Young for communication with the creators.
% AAAI is especially grateful to Peter Patel Schneider for

\bibliography{aaai22}

\begin{thebibliography}{28}
\providecommand{\natexlab}[1]{#1}

\bibitem[{Adiwardana et~al.(2020)Adiwardana, Luong, So, Hall, Fiedel,
  Thoppilan, Yang, Kulshreshtha, Nemade, Lu et~al.}]{adiwardana2020towards}
Adiwardana, D.; Luong, M.-T.; So, D.~R.; Hall, J.; Fiedel, N.; Thoppilan, R.;
  Yang, Z.; Kulshreshtha, A.; Nemade, G.; Lu, Y.; et~al. 2020.
\newblock Towards a human-like open-domain chatbot.
\newblock \emph{arXiv preprint arXiv:2001.09977}.

\bibitem[{Budzianowski et~al.(2018)Budzianowski, Wen, Tseng, Casanueva, Ultes,
  Ramadan, and Ga{\v{s}}i{\'c}}]{budzianowski2018multiwoz}
Budzianowski, P.; Wen, T.-H.; Tseng, B.-H.; Casanueva, I.; Ultes, S.; Ramadan,
  O.; and Ga{\v{s}}i{\'c}, M. 2018.
\newblock MultiWOZ--A Large-Scale Multi-Domain Wizard-of-Oz Dataset for
  Task-Oriented Dialogue Modelling.
\newblock \emph{arXiv preprint arXiv:1810.00278}.

\bibitem[{Byrne et~al.(2019)Byrne, Krishnamoorthi, Sankar, Neelakantan,
  Duckworth, Yavuz, Goodrich, Dubey, Cedilnik, and Kim}]{byrne2019taskmaster}
Byrne, B.; Krishnamoorthi, K.; Sankar, C.; Neelakantan, A.; Duckworth, D.;
  Yavuz, S.; Goodrich, B.; Dubey, A.; Cedilnik, A.; and Kim, K.-Y. 2019.
\newblock Taskmaster-1: Toward a realistic and diverse dialog dataset.
\newblock \emph{arXiv preprint arXiv:1909.05358}.

\bibitem[{Devlin et~al.(2018)Devlin, Chang, Lee, and
  Toutanova}]{devlin2018bert}
Devlin, J.; Chang, M.-W.; Lee, K.; and Toutanova, K. 2018.
\newblock Bert: Pre-training of deep bidirectional transformers for language
  understanding.
\newblock \emph{arXiv preprint arXiv:1810.04805}.

\bibitem[{Dinan et~al.(2018)Dinan, Roller, Shuster, Fan, Auli, and
  Weston}]{dinan2018wizard}
Dinan, E.; Roller, S.; Shuster, K.; Fan, A.; Auli, M.; and Weston, J. 2018.
\newblock Wizard of wikipedia: Knowledge-powered conversational agents.
\newblock \emph{arXiv preprint arXiv:1811.01241}.

\bibitem[{Ham et~al.(2020)Ham, Lee, Jang, and Kim}]{ham2020end}
Ham, D.; Lee, J.-G.; Jang, Y.; and Kim, K.-E. 2020.
\newblock End-to-End Neural Pipeline for Goal-Oriented Dialogue Systems using
  GPT-2.
\newblock ACL.

\bibitem[{Hosseini-Asl et~al.(2020)Hosseini-Asl, McCann, Wu, Yavuz, and
  Socher}]{hosseini2020simple}
Hosseini-Asl, E.; McCann, B.; Wu, C.-S.; Yavuz, S.; and Socher, R. 2020.
\newblock A simple language model for task-oriented dialogue.
\newblock \emph{arXiv preprint arXiv:2005.00796}.

\bibitem[{Houlsby et~al.(2019)Houlsby, Giurgiu, Jastrzebski, Morrone,
  De~Laroussilhe, Gesmundo, Attariyan, and Gelly}]{houlsby2019parameter}
Houlsby, N.; Giurgiu, A.; Jastrzebski, S.; Morrone, B.; De~Laroussilhe, Q.;
  Gesmundo, A.; Attariyan, M.; and Gelly, S. 2019.
\newblock Parameter-efficient transfer learning for NLP.
\newblock In \emph{International Conference on Machine Learning}, 2790--2799.
  PMLR.

\bibitem[{Howard and Cambria(2013)}]{howint}
Howard, N.; and Cambria, E. 2013.
\newblock Intention awareness: Improving upon situation awareness in
  human-centric environments.
\newblock \emph{Human-centric Computing and Information Sciences}, 3(9).

\bibitem[{Kim et~al.(2020)Kim, Eric, Gopalakrishnan, Hedayatnia, Liu, and
  Hakkani-Tur}]{kim2020beyond}
Kim, S.; Eric, M.; Gopalakrishnan, K.; Hedayatnia, B.; Liu, Y.; and
  Hakkani-Tur, D. 2020.
\newblock Beyond domain apis: Task-oriented conversational modeling with
  unstructured knowledge access.
\newblock \emph{arXiv preprint arXiv:2006.03533}.

\bibitem[{Li et~al.(2016)Li, Galley, Brockett, Gao, and
  Dolan}]{li2015diversityMMI}
Li, J.; Galley, M.; Brockett, C.; Gao, J.; and Dolan, B. 2016.
\newblock A Diversity-Promoting Objective Function for Neural Conversation
  Models.
\newblock In \emph{NAACL}, 110--119.

\bibitem[{Ma et~al.(2020)Ma, Nguyen, Xing, and Cambria}]{maasur}
Ma, Y.; Nguyen, K.~L.; Xing, F.; and Cambria, E. 2020.
\newblock A Survey on Empathetic Dialogue Systems.
\newblock \emph{Information Fusion}, 64: 50--70.

\bibitem[{Madotto et~al.(2020{\natexlab{a}})Madotto, Lin, Bang, and
  Fung}]{madotto2020adapter}
Madotto, A.; Lin, Z.; Bang, Y.; and Fung, P. 2020{\natexlab{a}}.
\newblock The Adapter-Bot: All-In-One Controllable Conversational Model.
\newblock \emph{arXiv preprint arXiv:2008.12579}.

\bibitem[{Madotto et~al.(2020{\natexlab{b}})Madotto, Lin, Wu, Shin, and
  Fung}]{madotto2020attention}
Madotto, A.; Lin, Z.; Wu, C.-S.; Shin, J.; and Fung, P. 2020{\natexlab{b}}.
\newblock Attention over Parameters for Dialogue Systems.
\newblock \emph{arXiv preprint arXiv:2001.01871}.

\bibitem[{Ni et~al.(2021)Ni, Young, Pandelea, Xue, Adiga, and
  Cambria}]{ni2021recent}
Ni, J.; Young, T.; Pandelea, V.; Xue, F.; Adiga, V.; and Cambria, E. 2021.
\newblock Recent Advances in Deep Learning Based Dialogue Systems: A Systematic
  Survey.
\newblock \emph{arXiv preprint arXiv:2105.04387}.

\bibitem[{Quan et~al.(2020)Quan, Zhang, Cao, Li, and Xiong}]{quan2020risawoz}
Quan, J.; Zhang, S.; Cao, Q.; Li, Z.; and Xiong, D. 2020.
\newblock RiSAWOZ: A Large-Scale Multi-Domain Wizard-of-Oz Dataset with Rich
  Semantic Annotations for Task-Oriented Dialogue Modeling.
\newblock In \emph{Proceedings of the 2020 Conference on Empirical Methods in
  Natural Language Processing (EMNLP)}, 930--940.

\bibitem[{Rastogi et~al.(2020)Rastogi, Zang, Sunkara, Gupta, and
  Khaitan}]{rastogi2020towards}
Rastogi, A.; Zang, X.; Sunkara, S.; Gupta, R.; and Khaitan, P. 2020.
\newblock Towards scalable multi-domain conversational agents: The
  schema-guided dialogue dataset.
\newblock In \emph{Proceedings of the AAAI Conference on Artificial
  Intelligence}, volume~34, 8689--8696.

\bibitem[{Roller et~al.(2020)Roller, Dinan, Goyal, Ju, Williamson, Liu, Xu,
  Ott, Shuster, Smith et~al.}]{roller2020recipes}
Roller, S.; Dinan, E.; Goyal, N.; Ju, D.; Williamson, M.; Liu, Y.; Xu, J.; Ott,
  M.; Shuster, K.; Smith, E.~M.; et~al. 2020.
\newblock Recipes for building an open-domain chatbot.
\newblock \emph{arXiv preprint arXiv:2004.13637}.

\bibitem[{Shuster et~al.(2019)Shuster, Ju, Roller, Dinan, Boureau, and
  Weston}]{shuster2019dialogue}
Shuster, K.; Ju, D.; Roller, S.; Dinan, E.; Boureau, Y.-L.; and Weston, J.
  2019.
\newblock The dialogue dodecathlon: Open-domain knowledge and image grounded
  conversational agents.
\newblock \emph{arXiv preprint arXiv:1911.03768}.

\bibitem[{Smith et~al.(2020)Smith, Williamson, Shuster, Weston, and
  Boureau}]{smith2020can}
Smith, E.~M.; Williamson, M.; Shuster, K.; Weston, J.; and Boureau, Y.-L. 2020.
\newblock Can You Put it All Together: Evaluating Conversational Agents'
  Ability to Blend Skills.
\newblock In \emph{Proceedings of the 58th Annual Meeting of the Association
  for Computational Linguistics}, 2021--2030.

\bibitem[{Sun et~al.(2020)Sun, Moon, Crook, Roller, Silvert, Liu, Wang, Liu,
  Cho, and Cardie}]{sun2020adding}
Sun, K.; Moon, S.; Crook, P.; Roller, S.; Silvert, B.; Liu, B.; Wang, Z.; Liu,
  H.; Cho, E.; and Cardie, C. 2020.
\newblock Adding Chit-Chats to Enhance Task-Oriented Dialogues.
\newblock \emph{arXiv preprint arXiv:2010.12757}.

\bibitem[{Sutskever, Vinyals, and Le(2014)}]{sutskever2014sequence}
Sutskever, I.; Vinyals, O.; and Le, Q.~V. 2014.
\newblock Sequence to sequence learning with neural networks.
\newblock In \emph{NIPS}, 3104--3112.

\bibitem[{Vinyals and Le(2015)}]{Vinyals2015A}
Vinyals, O.; and Le, Q. 2015.
\newblock A neural conversational model.
\newblock \emph{CoRR}, abs/1506.05869.

\bibitem[{Xu et~al.(2020)Xu, Peng, Xie, Cambria, Zhou, and Zheng}]{xuuend}
Xu, H.; Peng, H.; Xie, H.; Cambria, E.; Zhou, L.; and Zheng, W. 2020.
\newblock End-to-End latent-variable task-oriented dialogue system with exact
  log-likelihood optimization.
\newblock \emph{World Wide Web}, 23: 1989--2002.

\bibitem[{Ye, Manotumruksa, and Yilmaz(2021)}]{ye2021multiwoz}
Ye, F.; Manotumruksa, J.; and Yilmaz, E. 2021.
\newblock MultiWOZ 2.4: A Multi-Domain Task-Oriented Dialogue Dataset with
  Essential Annotation Corrections to Improve State Tracking Evaluation.
\newblock \emph{arXiv preprint arXiv:2104.00773}.

\bibitem[{Zhang et~al.(2018)Zhang, Dinan, Urbanek, Szlam, Kiela, and
  Weston}]{zhang2018personalizing}
Zhang, S.; Dinan, E.; Urbanek, J.; Szlam, A.; Kiela, D.; and Weston, J. 2018.
\newblock Personalizing dialogue agents: I have a dog, do you have pets too?
\newblock \emph{arXiv preprint arXiv:1801.07243}.

\bibitem[{Zhang et~al.(2019)Zhang, Sun, Galley, Chen, Brockett, Gao, Gao, Liu,
  and Dolan}]{zhang2019dialogpt}
Zhang, Y.; Sun, S.; Galley, M.; Chen, Y.-C.; Brockett, C.; Gao, X.; Gao, J.;
  Liu, J.; and Dolan, B. 2019.
\newblock Dialogpt: Large-scale generative pre-training for conversational
  response generation.
\newblock \emph{arXiv preprint arXiv:1911.00536}.

\bibitem[{Zhao et~al.(2017)Zhao, Lu, Lee, and Eskenazi}]{zhao2017generative}
Zhao, T.; Lu, A.; Lee, K.; and Eskenazi, M. 2017.
\newblock Generative encoder-decoder models for task-oriented spoken dialog
  systems with chatting capability.
\newblock \emph{arXiv preprint arXiv:1706.08476}.

\end{thebibliography}

\end{document}